\documentclass{IEEEtaes}

\usepackage{color,array,amsthm}
\usepackage{graphicx,xcolor}
\usepackage{booktabs}
\usepackage{amsmath,amssymb,amsfonts}
\usepackage{multirow}
\usepackage{pifont}

\jvol{XX}
\jnum{XX}
\jmonth{XXXXX}
\paper{1234567}
\pubyear{2022}
\doiinfo{TAES.2022.Doi Number}

\begin{document}

\title{Unified Restoration-Perception Learning: Maritime Infrared-Visible Image Fusion and Segmentation}

\author{Weichao Cai}
\affil{Xiamen University, Xiamen, Fujian 361005, China.} 

\author{Weiliang Huang}
\affil{University of Macau, Macau, China} 

\author{Biao Xue}
\affil{Xiamen University, Xiamen, Fujian 361005, China.} 

\author{Chao Huang}
\member{Member, IEEE}
\affil{Shenzhen Campus of Sun Yat-sen University, Shenzhen 518107, China}

\author{Fei Yuan}
\affil{Xiamen University, Xiamen, Fujian 361005, China.}

\author{Bob Zhang}
\member{Senior Member, IEEE}
\affil{University of Macau, Macau, China} 

\authoraddress{
    Weichao Cai, Biao Xue, and Fei Yuan are with the Laboratory of Underwater Acoustic Communication and Marine Information Technology, Ministry of Education, Xiamen University, Xiamen, Fujian 361005, China. (e-mail: caiweichao0914@stu.xmu.edu.cn, xuebiao@xmu.edu.cn, yuanfei@xmu.edu.cn).
    Weiliang Huang and Bob Zhang are with the PAMI Research Group, Department of Computer and Information Science, University of Macau, Macau, China (e-mail:yc47492@um.edu.mo, bobzhang@um.edu.mo).
    Chao Huang is with the School of Cyber Science and Technology, Shenzhen Campus of Sun Yat-sen University, Shenzhen 518107, China     (e-mail: huangch253@mail.sysu.edu.cn).
    Corresponding Author: Fei Yuan, Biao Xue.
}

\markboth{AUTHOR ET AL.}{SHORT ARTICLE TITLE}
\maketitle

\begin{abstract}
Marine scene understanding and segmentation plays a vital role in maritime monitoring and navigation safety. However, prevalent factors like fog and strong reflections in maritime environments cause severe image degradation, significantly compromising the stability of semantic perception. Existing restoration and enhancement methods typically target specific degradations or focus solely on visual quality, lacking end-to-end collaborative mechanisms that simultaneously improve structural recovery and semantic effectiveness. Moreover, publicly available infrared-visible datasets are predominantly collected from urban scenes, failing to capture the authentic characteristics of coupled degradations in marine environments. To address these challenges, the Infrared-Visible Maritime Ship Dataset (IVMSD) is proposed to cover various maritime scenarios under diverse weather and illumination conditions. Building upon this dataset, a Multi-task Complementary Learning Framework (MCLF) is proposed to collaboratively perform image restoration, multimodal fusion, and semantic segmentation within a unified architecture. The framework includes a Frequency-Spatial Enhancement Complementary (FSEC) module for degradation suppression and structural enhancement, a Semantic-Visual Consistency Attention (SVCA) module for semantic-consistent guidance, and a cross-modality guided attention mechanism for selective fusion. 
Experimental results on IVMSD demonstrate that the proposed method achieves state-of-the-art segmentation performance, significantly enhancing robustness and perceptual quality under complex maritime conditions.
\end{abstract}


\section{INTRODUCTION}
\label{sec:intro}

\begin{figure}[h!]
    \includegraphics[width=\linewidth]{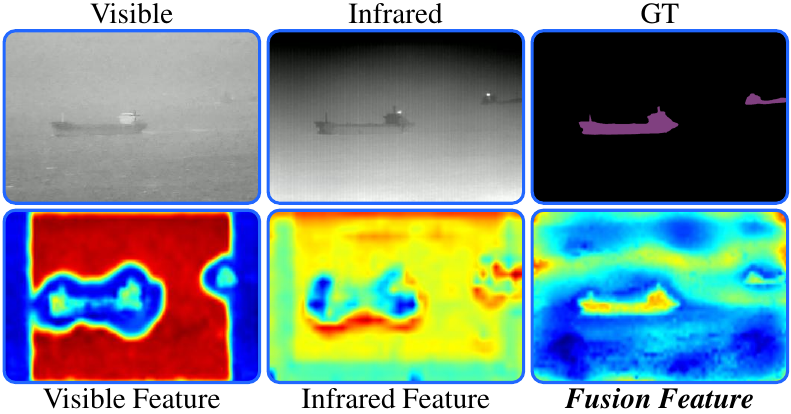}
    \caption{Under complex maritime conditions, single-modality features exhibit limited discriminability due to severe degradation and modality-specific information loss. In contrast, the proposed method explicitly models degradation patterns and performs multimodal feature fusion, thereby substantially enhancing the robustness and discriminative capability of the learned representations.}
    \label{fig:motivation}
\end{figure}

Maritime defense and ocean surveillance require reliable perception of vessels and surrounding environments in complex marine scenarios \cite{10153477,10693559}. To support these tasks, remote sensing technologies are widely used to acquire maritime scene data over wide spatial ranges. For downstream maritime applications, the key challenge lies not only in acquiring scene data but also in interpreting the semantic information of maritime scenes, including target categories, spatial regions, and environmental context.
From this perspective, marine scene understanding has become an essential problem in intelligent maritime perception.
As this process relies heavily on accurate environmental perception, optical sensors, with their high spatial resolution and rich textural features, have become an important tool \cite{11328754,10024907}. However, the perceptual capability of a single modality has obvious limitations, as shown in \autoref{fig:motivation}: 1) visible light images provide abundant detail but suffer from significant performance degradation under low-light, haze, or strong reflection conditions; 2) infrared images offer robust thermal radiation imaging characteristics, but lack sufficient texture and color details.

To overcome the limitations of single modality, Infrared–Visible Multi-modal Fusion (IVMF) has been extensively studied. By combining the complementary features of the two modalities, it achieves more robust scene representation and has shown significant potential in fields such as autonomous driving \cite{ha2017mfnet,franchi2024infraparis}, intelligent surveillance \cite{zhang2024multi}, and remote sensing \cite{nie2023cross,ma2024visible}. However, in complex marine conditions, environmental interference (such as sea fog and thermal noise) severely degrades image quality and jeopardizes the robustness of fusion models, thus limiting their practical reliability.
To address these issues, some studies \cite{cho2021rethinking, li2024focus, li2021different, qin2020ffa} have proposed various image enhancement and restoration methods to improve visual quality and mitigate degradation effects. However, most existing methods are developed to address specific, individual types of degradation and struggle to handle multi-source degradations.

Recently emerging all-in-one enhancement networks \cite{yu2024multi, cao2025mmaif, luo2023controlling} attempt to handle multiple degradations within a unified framework, but their optimization objectives primarily focus on visual quality instead of downstream task performance. More fundamentally, existing methods generally lack an end-to-end optimization mechanism, making it difficult to achieve synergistic improvement between low-level visual restoration and high-level semantic understanding. This contradiction constitutes a key bottleneck in current multi-modal marine scene perception.
Meanwhile, the lack of infrared–visible maritime datasets containing multiple degraded types is another critical factor severely hindering the development of this field. Existing publicly available infrared-visible datasets are mostly collected in urban environments and fail to reflect the complex conditions prevalent in marine scenes. 


To address these issues, this paper constructed an Infrared–Visible Maritime Ship Dataset (IVMSD). This dataset was collected in real maritime environments, covering a variety of complex weather and lighting conditions. The data primarily focuses on ship scenes, containing ship targets of different sizes, poses, and motion states. This dataset provides a realistic benchmark for evaluating multimodal perception performance under complex sea conditions and offers reliable data support for researching the end-to-end fusion–segmentation collaborative optimization framework. Based on this dataset, a unified complementary learning framework is proposed to achieve collaborative optimization of image restoration, multimodal fusion, and semantic segmentation tasks under an end-to-end unified architecture. Specifically, a Frequency–Spatial Complementary Enhancement (FSCE) module is designed to handle multiple types of degradation. This mechanism integrates frequency-domain and spatial-domain features, suppressing degradations such as blur and noise at the spectral level, while enhancing structural information at the spatial level. Subsequently, a Semantic–Visual Consistency Attention (SVCA) module is proposed, which leverages semantic priors to inversely guide the visual enhancement, thereby bridging the gap between low-level restoration and high-level semantic tasks. This dual-level collaborative strategy significantly improves feature stability and task relevance under complex degradation conditions. 
Additionally, a cross-modal guided attention mechanism is introduced, designed to selectively fuse semantically relevant features by gated cross-scale attention, thereby suppressing redundant information and noise propagation.This effectively integrates the complementary advantages of infrared and visible modalities, enhancing the discriminability and consistency of multimodal features.

Therefore, the contributions of this paper can be summarized as follows.

\textit{1)} Dataset Construction: A realistic and high-quality Infrared-Visible Maritime Ship Dataset (IVMSD) is constructed, which covers multi-modal ship targets under complex maritime weather conditions and includes diverse variations in target scale, orientation, and motion. IVMD fills the gap of lacking real maritime multi-degradation datasets.

\textit{2)} Framework Innovation: A Multi-task Complementary Learning Framework (MCLF) is proposed that realizes unified restoration-perception learning. This framework bridges the gap between low-level degradation restoration and high-level semantic perception, achieving collaborative optimization of the two tasks.

\textit{3)} Module Design: MCLF integrates two key modules that the Frequency-Spatial Enhancement Complementary  (FSEC) module suppresses multi-type degradations via frequency-spatial joint optimization and the Semantic-Visual Consistency Attention (SVCA) module ensures semantic consistency of multi-modal features, enhancing perceptual robustness.

\textit{4)} Experimental Validation: Extensive experiments on IVMSD demonstrate that the proposed method outperforms state-of-the-art baselines, verifying the effectiveness of the unified restoration-perception learning paradigm and the superiority of the proposed modules in complex maritime scenarios.

\section{RELATED WORKS}
\label{sec:formatting}

\subsection{Understanding and Segmentation Methods for Maritime Scenarios}


With the advent of maritime-specific datasets such as SeaShips \cite{shao2018seaships}, MODD2 \cite{bovcon2018stereo}, and MID \cite{liu2021efficient}, which include region-level or pixel-level annotations and diverse scenarios, the research paradigm has undergone a pivotal shift. It has moved from unsupervised learning \cite{zhang2023two, vandaele2020automated} to fully-supervised, end-to-end model training. In this context, a growing number of supervised learning methods \cite{shao2019saliency, xu2021hierarchical, zhang2024infrared} tailored for complex maritime environments have been proposed. These methods have effectively addressed key tasks including ship detection, coastline semantic segmentation, waterway obstacle perception, and dynamic maritime scene understanding, significantly enhancing the intelligence of maritime monitoring, autonomous navigation, and voyage safety systems. For example, IMODN \cite{zhang2024infrared} proposes an infrared maritime object detection network with feature enhancement and adjacent fusion to improve the detection of small and dim targets.

However, existing datasets still exhibit significant limitations in terms of modality, resolution, scene complexity, and annotation granularity. For instance, SeaShips \cite{shao2018seaships} lacks the infrared modality; MODD2 \cite{bovcon2018stereo} suffers from low resolution and relatively simplistic scenes; and MID \cite{liu2021efficient} lacks fine-grained, pixel-level segmentation annotations for obstacles. This lag in data capability severely constrains models from achieving precise perception and understanding of multi-scale, small, and blurry targets in complex maritime environments.


\subsection{Restoration Methods for Degraded Images}


The rapid development of deep learning has given rise to numerous effective methods \cite{zhang2017beyond, jiang2020multi, zhang2024depth, chen2021all} for image restoration tasks, including deblurring, dehazing, deraining, and desnowing. While these advancements have significantly boosted performance on individual tasks, handling diverse types of image degradations within a unified framework presents new challenges, as each degradation type possesses distinct characteristics that require tailored network designs and optimization strategies. 

To improve the general applicability of image restoration methods, some studies \cite{yu2024multi, cao2025mmaif, liu2025uhd, yin2025dformerv2} have begun to explore how to address multiple heterogeneous degradations by unified frameworks. For example, MEAS \cite{yu2024multi} proposes a multi-expert adaptive selection mechanism that dynamically selects and integrates different expert networks to achieve all-in-one multi-task image restoration. These unified frameworks have shown promise in handling heterogeneous degradations, but they focus primarily on enhancing the visual quality of the restored images, while overlooking the impact of the restoration results on the performance of downstream vision tasks. This limitation may result in restored images that perform well in subjective quality assessments, yet fail to deliver the expected performance gains in practical application systems, thereby limiting their practical value in real-world scenarios.


\section{Infrared-Visible Maritime Ship Dataset}
Since existing maritime datasets exhibit various limitations, they hardly meet the requirements for rigorous and comprehensive evaluation of multimodal perception models under complex degradation conditions. Therefore, the primary objective is to construct a high-resolution, multimodal comprehensive maritime dataset covering complex marine environmental conditions, designed to evaluate the integrated performance of multimodal fusion and semantic segmentation algorithms under realistic maritime conditions. It simultaneously provides high-resolution infrared-visible image pairs under diverse complex maritime scenarios and pixel-level segmentation annotations. This benchmark aims to establish a unified and challenging real-world platform for collaborative research and performance evaluation in multimodal image restoration, fusion, and semantic segmentation tasks. First, the data acquisition and preprocessing methodologies are elaborated. Subsequently, an efficient semi-automatic annotation process and dataset configuration are described.

\subsection{Dataset Constructions}

\textit{1) Data Collection:} To enhance the robustness of maritime perception models in real-world complex environments, this work conducted data collection using shore-based platforms in a Chinese coastal area under various adverse weather conditions. It deployed an integrated infrared-visible acquisition system, which incorporates high-performance thermal infrared cameras and visible-light cameras.  Equipped with a synchronization control unit, the system enables hardware-synchronized triggering for simultaneous data capture, ensuring strict spatiotemporal alignment between the two modalities. The inherent stability of shore-based platforms guarantees prolonged, uninterrupted data acquisition, particularly advantageous for collecting maritime scene data under diverse weather conditions. During this phase, this paper obtains visible images with a resolution of $1520 \times 2688$ and infrared images with a resolution of $1024 \times 1280$.

\textit{2) Data preprocessing:} This paper first filtered the raw data, retaining valid sequences containing various types of ships under diverse weather conditions. Subsequently, this paper performed automatic registration using MINIMA \cite{ren2025minima} to achieve pixel-level alignment between visible and infrared images. MINIMA \cite{ren2025minima} is a  universal image matching model, capable of effectively achieving pixel-level registration across multiple modalities such as visible, infrared, and depth data. Finally, to standardize input dimensions and focus on key maritime targets, this paper cropped all data to images with an average size of $720 \times 1280$. Through this preprocessing pipeline, this paper ultimately obtained a dataset containing 2,313 rigorously aligned high-quality infrared-visible image pairs. This dataset notably includes a substantial number of samples under complex conditions such as limited visibility, poor illumination, and noise interference, providing crucial data support for studying model perception performance in extreme scenarios.

\textit{3) Data annotation:} To construct a high-quality pixel-level annotated dataset, this paper adopts a semi-automatic annotation pipeline that integrates cutting-edge visual foundational models with manual verification. This workflow aims to ensure annotation accuracy, while significantly improving labeling efficiency. The specific procedures are as follows:

\begin{figure}
    \centering
    \includegraphics[width=\linewidth]{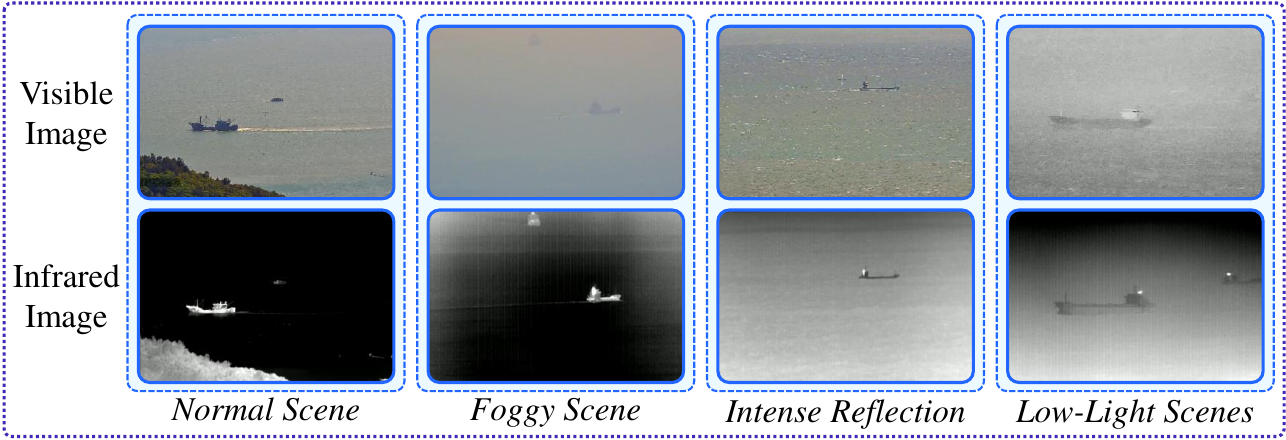}
        \caption{Examples of different scenes included in the IVMSD dataset.}
    \label{fig:scene}
\end{figure}

\textit{Interactive Annotation of Initial Frames}: For each video sequence, this paper first leverage the powerful  visual foundational model, Segment Anything Model 2 (SAM2 \cite{ravi2024sam2}) , for initial interaction. By providing minimal manual prompts (such as object bounding boxes or keypoints) on the first frame, SAM2 generates high-quality pixel-level segmentation masks, enabling rapid and accurate annotation of all target objects in the initial frame.

\textit{Video Object Segmentation Propagation}: Using the precise annotations from the first frame as reference, this paper subsequently apply SAM2's video segmentation capability to automatically track the key objects across all subsequent frames. The model effectively utilizes temporal information to robustly track object motion  throughout the video, thereby generating coherent and complete frame-by-frame object mask sequences.

\textit{Manual Refinement and Correction}: Given the challenges fully automated segmentation faces in dynamic and complex maritime environments, this paper introduces a crucial manual refinement and correction phase. Annotation experts systematically review the automatically generated mask sequences, focusing particularly on segmentation ambiguities and errors caused by maritime-specific complexities. Otherwise, it will include some issues such as missing small targets and insufficient boundary precision. 

In summary, while fully leveraging the strong generalization capability of foundation models to enhance efficiency, the incorporation of expert knowledge through manual intervention effectively ensures the accuracy and reliability of the final annotations.

\subsection{Dataset Description}
Despite significant advances in maritime perception research, existing public datasets still exhibit notable limitations in modality completeness, scene complexity, and annotation quality. The scarcity of high-quality, multimodal maritime datasets severely constrains the development of robust perception models capable of adapting to real-world complex marine conditions. To address this, this dataset was systematically designed with careful planning for scene diversity and target category balance during the design phase. This section will elaborate on the dataset's configuration, including the analysis of scenes and maritime target characteristics.

As shown in \autoref{fig:scene}, the dataset comprises rigorously aligned infrared-visible image pairs, categorized according to key challenges in practical maritime applications into the following scene types: Normal Scenes, Foggy Scenes, Low-Light Scenes, and Intense Reflection Scenes. 

\textit{Normal Scenes}: Characterized by sufficient illumination, high visibility, and calm sea conditions. Visible images retain rich color and texture information, while infrared images clearly display target thermal distributions based on thermal inertia. The two modalities complement each other, serving as benchmark conditions for model performance evaluation.

\textit{Foggy Scenes}: Presence of suspended water droplets in the atmosphere causes significant degradation in overall contrast and saturation in visible images, with blurred target edges. The infrared spectrum exhibits certain penetration capability through haze, typically maintaining more complete target contours and structures in foggy conditions, effectively mitigating perception challenges caused by limited visibility.

\textit{Low-Light Scenes}: Weakened ambient lighting results in low signal-to-noise ratios and severe detail loss in visible images. In such scenarios, the infrared modality relies entirely on targets' inherent thermal radiation characteristics for imaging, independent of ambient lighting. Thermal contrast becomes the crucial perceptual cue, particularly for vessel targets with heat sources like engines.

\textit{Intense Reflection Scenes}: Under sunlight and wave undulations, the sea surface produces dynamic, strong specular reflections. In visible images, these reflections cause local overexposure and glare, degrading target texture, color, and contour information. In contrast, infrared imaging is based on thermal radiation and is less sensitive to reflected light, allowing clearer target thermal distributions and shapes and thus more stable perception in high-reflection environments.

\subsection{Dataset Characteristics}
The dataset comprises four selected categories of civilian maritime targets: cargo ships, fishing boats, sand dredgers, and speedboats. This selection systematically covers the core target types in maritime scenes, presenting a representative set of challenges in terms of size, speed, structure, and thermodynamic characteristics.

\textit{1) Cargo ships:} As large, slow-moving targets, cargo ships are valuable for evaluating model segmentation accuracy on major structures and the ability to identify massive heat sources (e.g., engines and hulls) under harsh conditions.

\textit{2) Fishing boats:} Fishing boats represent typical small-to-medium-sized targets, exhibit drastic attitude variations, and show weak thermal signatures. They primarily challenge model robustness in detecting small targets under low signal-to-noise ratios.

\textit{3) Sand dredgers:} As specialized working vessels, sand dredgers possess unique mechanical structures and generate specific thermal distribution patterns during operations. They provide critical samples for recognizing task-specific behavioral patterns.

\textit{4) Speedboats:} It represents high-speed small target, exhibiting significant motion blur and concentrated high-temperature heat sources. They test model capabilities in moving-target segmentation and rapid cross-modal association.

These four target categories demonstrate notable complementary and divergent characteristics across visible and infrared modalities, which ensures the dataset can comprehensively and fairly evaluate the integrated performance of multimodal perception models in realistic complex maritime environments.

\begin{figure*}
    \centering
    \includegraphics[width=\linewidth]{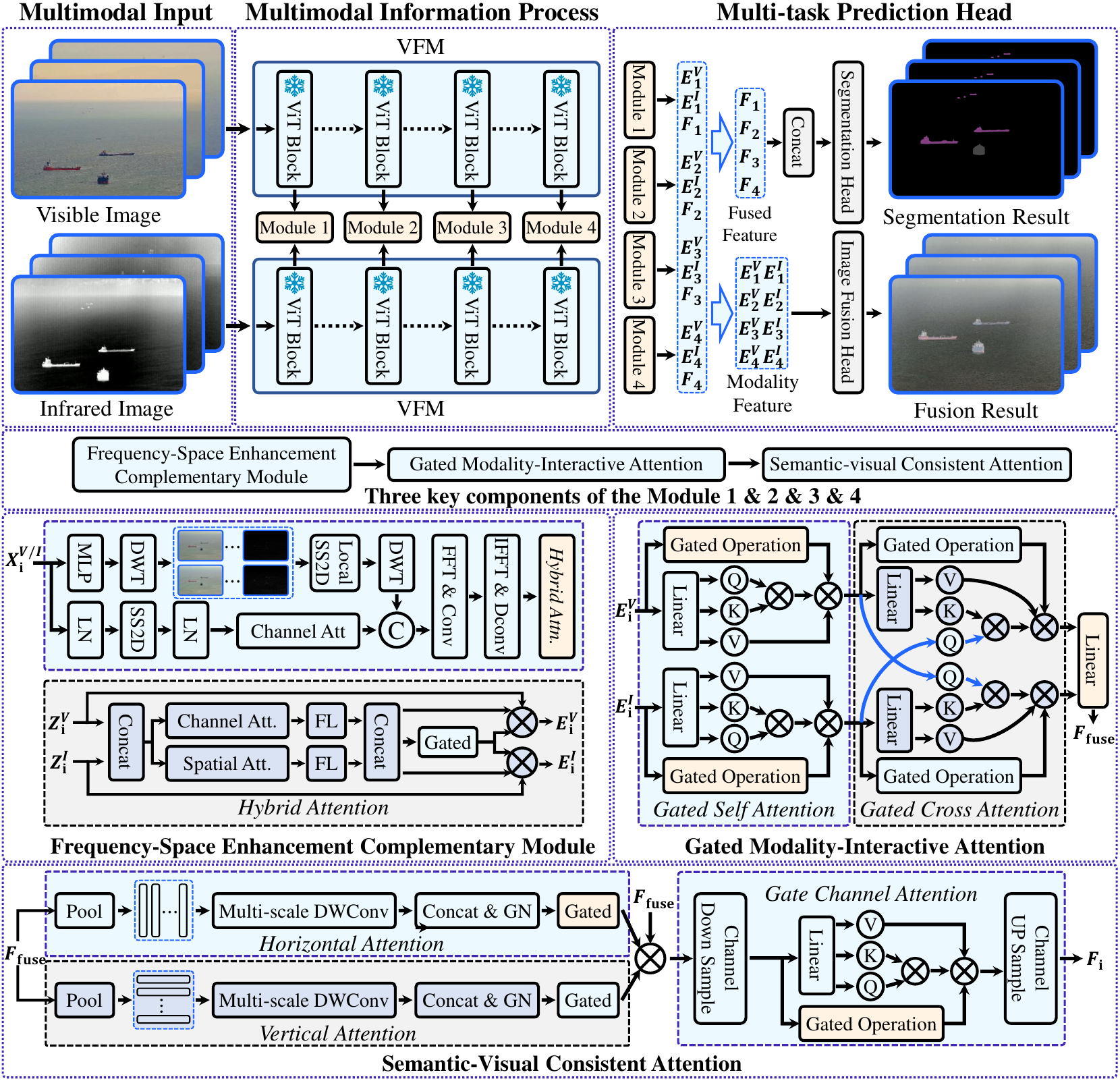}
    \caption{The overview of the proposed Multi-task Complementary Learning Framework.}
    \label{fig:Framework}
    \vspace{-2mm}
\end{figure*}

\section{Method}
\subsection{Multi-task Complementary Learning Framework}
Based on the IVMSD benchmark, this paper proposes a Multi-task Complementary Learning Framework (MCLF), as illustrated in \autoref{fig:Framework}.
Unlike conventional cascaded designs that separate enhancement and perception, MCLF formulates fusion and segmentation as semantically driven enhancement tasks.
Through complementary integration of infrared and visible features under semantic supervision, MCLF suppresses degradations, enhancing structural information and contrast. The enhanced features improve semantic discriminability, while semantic information guides the enhancement process.
As illustrated in \autoref{fig:Framework}, this paper uniformly sample four layers of outputs from low-level to high-level features to serve enhancement and fusion, which contain varying degrees of low-level structural information and high-level semantic information respectively. These intermediate features are then passed into key modules for feature enhancement targeting multiple degradations. Finally, the obtained fused features and modality features are utilized for segmentation and fusion tasks, respectively.

The MCLF consists of three key modules:

\textit{1) Frequency-Spatial Enhancement Complementary (FSEC):} Fuses frequency- and spatial-domain representations to suppress degradations, enhancing structural information spatially, effectively handling diverse and coupled degradations.

\textit{2) Gated Modality-Interactive Attention (GMIA):} Employs gated cross-attention to selectively integrate multimodal features, establishing semantically consistent interactions across modalities, suppressing redundancy and noise propagation, thus improving feature discriminability and consistency.

\textit{3) Semantic–Visual Consistency Attention (SVCA):} Exploits the intrinsic consistency between fusion and segmentation to strengthen semantic discriminability, enabling semantic cues to guide enhancement toward task-relevant regions for joint optimization of enhancement and perception.


\subsection{Frequency-Spatial Enhancement Complementary}
Under complex maritime conditions, images are affected by multiple types of degradation, such as fog, sea spray, strong reflections, and thermal noise. These degradations manifest differently across frequency bands, leading to texture blurring, structural distortion, and reduced contrast.
From the frequency-domain perspective, different degradations exhibit distinct spectral patterns.
Noise and sea-spray interference affect the high-frequency bands, disrupting edge and texture components and leading to loss of fine details.
In contrast, fog, strong reflections, and thermal noise degrade the low-frequency bands, causing global brightness shifts and contrast attenuation.
Consequently, maritime images under complex conditions suffer from both high-frequency detail suppression and low-frequency structural degradation, which jointly challenge feature modeling and multimodal fusion.

To address these issues, this paper proposes the FSEC module that jointly models degradation characteristics in both frequency and spatial domains. 
This module leverages the discrete wavelet transform (DWT) for multi-resolution decomposition to capture localized degradation patterns across sub-bands while marking degradation-affected targets in the spatial domain. 
The Fast Fourier Transform (FFT)-based filtering performs global frequency refinement for holistic optimization of frequency sub-bands impacted by diverse degradation patterns. Through the adaptive band adjustment strategy, it further achieves cross-band information compensation and structural enhancement.

Specifically, given a feature map $\mathbf{X}^{V/I} \in \mathbb{R}^{B \times C \times H \times W}$ from the encoder ($V/I$ denotes the visible or infrared, respectively), second-level DWT is adopted, which decomposes the feature into hierarchical sub-bands that disentangle low-frequency structural information and high-frequency texture details:
\begin{align}
    \{\mathbf{X}_{\text{LL}}^{(j)}, \mathbf{X}_{\text{LH}}^{(j)}, \mathbf{X}_{\text{HL}}^{(j)}, \mathbf{X}_{\text{HH}}^{(j)}\}_{j=1}^2 = \text{DWT}(\mathbf{X}),
\end{align}
where $j$ is the index of decomposition level, $X_{\text{LL}}$ represents the low-low frequency approximation, and $\{X_{\text{LH}}, X_{\text{HL}}, X_{\text{HH}}\}$ are high-frequency detail coefficients. This frequency-aware decomposition facilitates spatially adaptive modeling, as the subsequent attention module can exploit distinct frequency components to enhance discrimination:
Then these coefficients is reorganized into a compact tensor $\mathbf{H} \in \mathbb{R}^{B \times C \times H \times W}$ via spatial rearrangement $\mathcal{R}(\cdot)$:
    $\mathbf{H} = \mathcal{R}(\mathbf{X}_{\text{LL}}^{(2)}, \{\mathbf{X}_{\text{LH}}^{(j)}, \mathbf{X}_{\text{HL}}^{(j)}, \mathbf{X}_{\text{HH}}^{(j)}\}_{j=1}^2).$
This rearrangement preserves the complete frequency hierarchy within the spatial dimensions, allowing subsequent operations to exploit both local and global frequency contexts.
Then apply local selective scanning via local 2D Selective Scan (SS2D) \cite{SS2D_local} to capture fine-grained sub-band information. Local SS2D processes the feature map in non-overlapping windows of size $w \times h$ to capture fine-grained patterns. Thus, four scanning directions (row-major, column-major, and their flipped versions) is adopted, generating in a local state space sequence $\{h_k\}_{k=1}^4$. The enhanced local feature $\mathbf{H}$ is computed as:
$\mathbf{H'} = \mathbf{H} + \text{SS2D}_\text{local}(\text{LN}(\mathbf{H})),$
where $\text{LN}(\cdot)$ denotes Layer Normalization. The wavelet coefficients are then reconstructed via inverse DWT to restore spatial features:
\begin{align}
    \mathbf{\hat{X}} = \text{DWT}^{-1}(\mathbf{X}_{\text{LL}}^{(2)}, \{\mathbf{X}_{\text{LH}}^{(j)}, \mathbf{X}_{\text{HL}}^{(j)}, \mathbf{X}_{\text{HH}}^{(j)}\}_{j=1}^2),
\end{align}
where the coefficients are replaced with their enhanced versions from $\mathbf{H'}$.

Simultaneously, this paper enhance the original feature $\mathbf{X}$ using vanilla SS2D and channel attention to mark degradation-affected targets in the spatial domain. The output $\mathbf{X'}$ is obtained by: $\mathbf{X'} = \mathbf{X} + \text{SS2D}(\text{LN}(\mathbf{X})) + \text{CA}(\text{LN}(\mathbf{X}))$, where $\text{CA}$ denotes channel attention.

Subsequently, to synergistically optimize each frequency sub-band affected by different degradation patterns from the global perspective, Fast Fourier Transform (FFT) is adopted to the concatenated feature $[\mathbf{X'}, \mathbf{\hat{X}}]$:
\begin{align}
    \mathcal{F} = \mathcal{F}_\text{r}\{ [\mathbf{X'}, \mathbf{\hat{X}}] \} \in \mathbb{C}^{B \times 2C \times H \times \lfloor W/2+1 \rfloor},
\end{align}
where $\mathcal{F}_\text{r}$ denotes the 2D FFT. This paper decompose $\mathcal{F}$ into magnitude $|\mathcal{F}|$ and phase $\angle\mathcal{F}$ components. The magnitude spectrum is processed by a learnable denoising convolution $\mathcal{D}_\theta$: $|\mathcal{F}|_\text{enh} = \mathcal{D}_\theta(|\mathcal{F}|)$. The enhanced spectrum is then reconstructed by the IFFT $\mathcal{F}_\text{r}^{-1}$. A depthwise convolution is then applied to generate the enhanced feature $\mathbf{Y}$.

Furthermore, infrared and visible images possess complementary modality-specific information. To effectively leverage the synergy of this cross-modal information and achieve cross-modal complementary enhancement, a Hybrid Attention is adopted. Specifically, given two modality-specific features \(\mathbf{Y}^I, \mathbf{Y}^V \in \mathbb{R}^{B \times C \times H \times W}\), this Hybrid Attention computes channel attention weights $\mathbf{W}_c$ and spatial attention $\mathbf{W}_s$ by combining global average and max pooling statistics across concatenated feature $[\mathbf{Y}^I, \mathbf{Y}^V]$. It can highlight the key channels' information while simultaneously reinforcing discriminative spatial structural information. The hybrid attention map combines these two weights $\mathbf{M} = \mathbf{W}_c \odot \mathbf{W}_s \in \mathbb{R}^{2 \times B \times C \times H \times W}$. A learnable gating mechanism dynamically modulates the modality contributions $G = \sigma(\text{Linear}_{2C\to C}(\text{Concat}(F_i, F_v))).$, adaptively determining the enhancement strategy for the modalities feature. Finally, cross-modal complementarity is achieved through gating coefficients $\mathbf{E}^V = \mathbf{Y}^I + (1 - G) \odot M \odot \mathbf{Y}^V$ and $\mathbf{E}^I = \mathbf{Y}^V + (1 - G) \odot M \odot \mathbf{Y}^I$.

\subsection{Gated Modality-Interactive Attention}
Given the visible and infrared modal features, $\mathbf{E}^V$ and $\mathbf{E}^I$ respectively, this module acquires internal feature representation through a headwise gated self-attention unit for $\mathbf{E}^V$ and $\mathbf{E}^I$: $\hat{\mathbf{F}}^{V} = \text{Gate}(\sigma(\mathbf{Q} \mathbf{K}^{\top} / \sqrt{d}))\mathbf{V}$ and $\hat{\mathbf{F}}^{I} = \text{Gate}(\sigma(\mathbf{Q} \mathbf{K}^{\top} / \sqrt{d})\mathbf{V}$. Here, $\mathbf{Q}$, $\mathbf{K}$, and $\mathbf{V}$ respectively denote the Query, Key, and Value maps, and $\sigma(\cdot)$ represents the Softmax normalization operation. Furthermore, the $\text{Gate}(\cdot)$ function signifies the head-wise gating mechanism, which is employed to suppress noise-dominant channels and reinforce task-relevant features adaptively.

Visible and infrared images exhibit inconsistent degradation, vanilla cross-attention easily introduces invalid or even misleading features. To address this issue, the Gating Mechanism is introduced into the cross-attention process, performing Bidirectional Gated Cross-Attention between the two modalities.
The attention maps for the flow from infrared to visible-light ($\mathbf{A}^{{V} \leftarrow {I}}$) and visible to infrared ($\mathbf{A}^{{I} \leftarrow {V}}$) are computed as follows: 
$\{\mathbf{Q}^{{V}} = \mathbf{F}^{{V}}\mathbf{W}_Q, [\mathbf{K}^{{I}}, {V}^{{I}}] = \mathbf{F}^{{I}}\mathbf{W}_{KV}, \mathbf{A}^{{V} \leftarrow {I}} = \text{Attn}({\mathbf{Q}^{{V}}, (\mathbf{K}^{{I}})^{\top}})$\} 
and 
$\{\mathbf{Q}^{{I}} = \mathbf{F}^{{I}} \mathbf{W}_Q', [\mathbf{K}^{{V}}, {V}^{{V}}] = \mathbf{F}^{{V}} \mathbf{W}_{KV}', \mathbf{A}^{{I} \leftarrow {V}} = \text{Attn}({\mathbf{Q}^{{I}}, (\mathbf{K}^{{V}})^{\top}})\}$. 


The cross-modal information flow is then regulated by gating signals generated from the query modality, which achieves the selective fusion of semantically relevant features: $\{\mathbf{G}^{{V}} = \sigma(\mathbf{F}^{{V}}\mathbf{W}_g), \mathbf{F}^{{V}} = (\mathbf{A}^{{V} \leftarrow {I}}{V}^{{I}}) \odot \mathbf{G}^{{V}}\}$ 
and 
$\{\mathbf{G}^{{I}} = \sigma(\mathbf{F}^{{I}} \mathbf{W}_g'), \mathbf{F}^{{I}} = (\mathbf{A}^{{I} \leftarrow {v}} {v}^{{v}}) \odot \mathbf{G}^{{I}}\}$. 

Finally, the fused feature $\mathbf{F}_{\text{fuse}}$ is generated via a linear layer:
$\mathbf{F}_{\text{fuse}} = \psi(\mathbf{F}^I, \mathbf{F}^V)$, 
where $\psi(\cdot)$ denotes a Linear operation. $\mathbf{F}^I$ and $\mathbf{F}^V$ are the final gated feature representations from the infrared and visible-light branches, respectively.

\subsection{Semantic-visual Consistent Attention}
Considering the intrinsic consistency between multi-modal segmentation and fusion tasks \cite{zhang2024mrfs}, this paper aim to enhance the discriminability of semantic representations further. This allows the downstream task to utilize semantic cues to reverse guide the enhancement process, emphasizing task-relevant regions and achieving joint optimization between enhancement and downstream perception.
Thus, this paper proposes the SVCA module, which performs adaptive feature enhancement under multi-modal semantic constraints by integrating axis-decoupled spatial attention and gate-based channel self-attention mechanisms. In maritime scenes, targets exhibit strong directional and structural continuity while semantic regions present uneven geometric distributions along the height and width dimensions. 
To accommodate this property, this paper adopts an axis-decoupled strategy that decomposes the conventional 2D modeling into two independent 1D attention branches. This design enables the model to separately capture vertical and horizontal structural dependencies, thereby improving structure sensitivity and semantic response consistency.

Specifically, given the fused multi-modal feature $\mathbf{F}_{\text{fuse}}$, an axis-decoupled spatial attention mechanism is first employed to model structural dependencies along the height and width dimensions, respectively: $\mathbf{X}^h = \frac{1}{W}\sum_{w}\mathbf{X}_{\text{fuse}}, \mathbf{X}^w = \frac{1}{H}\sum_{h}\mathbf{X}_{\text{fuse}}$.
Each group's features are independently processed through depthwise separable 1D convolutions to capture multi-scale long-range dependencies, where the kernel sizes $k_i \in \{3,5,7,9\}$ increase progressively to characterize alignment patterns from local details to global structures:
$\mathbf{A}^h_i = \text{DWConv}{k_i}(\mathbf{X}^h_i), \mathbf{A}^w_i = \text{DWConv}{k_i}(\mathbf{X}^w_i)$. 
The responses of all groups are concatenated along the channel dimension and normalized via Group Normalization (GN). Then, a gating function 
$\sigma$ is applied to generate the attention maps, which effectively suppress cross-modal noise activations while highlighting task-relevant structural regions: $\mathbf{M}^h = \sigma_h(\text{GN}([\mathbf{A}^h_i])), \mathbf{M}^w = \sigma_w(\text{GN}([\mathbf{A}^w_i)])$, where $[\cdot]$ denotes channel concatenation. The spatially recalibrated features are efficiently reconstructed through tensor multiplication: $\mathbf{X}' = \mathbf{X}_{\text{fuse}} \odot \mathbf{M}^h\mathbf{M}^{w\top}$. Subsequently, a gate-based channel self-attention mechanism is introduced to adaptively suppress redundant channel responses under semantic feedback from the multi-modal fusion task. This design ensures that semantically salient regions receive higher weights in the feature space, thereby guiding the enhancement process to focus on task-relevant regions. The final output feature is denotes as $\mathbf{F}$. 

\subsection{Multi-task Head}
In the proposed Multi-task Complementary Learning Framework, enhancement, cross-modal interaction, and semantic–visual consistency jointly produce a set of rich multi-scale multimodal representations. To fully exploit the structural cues and semantic information encoded in these features, this paper introduces two complementary output heads: a Segmentation Head and a Fusion Head. 
Integrated under a unified multi-task scheme, these heads establish a bidirectional coupling mechanism between enhancement, fusion, and segmentation, forming a crucial closed-loop learning component of MRFS.

\textit{Segmentation Head:} The segmentation branch follows a cross-scale reconstruction paradigm using a lightweight DPT-style decoder. This head was designed with the following aspects: 1) Segmentation loss propagates back through FSC, GMIA, and SVCA, making the enhancement stage semantically aware and improving modality alignment; 2) Lower-level detail and higher-level semantics are jointly exploited, significantly improving boundary sharpness and small-object recognition under challenging maritime conditions.
Given fused features $\left\{ \mathbf{F}_1, \mathbf{F}_2, \mathbf{F}_3, \mathbf{F}_4 \right\}$,
token sequences are first reshaped into spatial maps and linearly projected:
    $\hat{\mathbf{F}}_i = \mathrm{Proj}_i(\mathrm{Reshape}(\mathbf{F}_i)), i=\{1,2,3,4\}$
Convolutional refinement aligns the statistics across levels:
    $\mathbf{Z}_i = \mathrm{Conv}_i(\hat{\mathbf{F}}_i).$
Upsampling brings all features to the highest spatial resolution:
    $\tilde{\mathbf{Z}}_i = \mathrm{Up}(\mathbf{Z}_i, H, W).$
The final segmentation result is obtained by aggregating all scales:
    $\mathbf{Y}_{\text{seg}} = \mathrm{Conv}_{1\times1}([\tilde{\mathbf{Z}}_1,\tilde{\mathbf{Z}}_2,\tilde{\mathbf{Z}}_3,\tilde{\mathbf{Z}}_4]).$    

\textit{Fusion Head:} Activated only during training, the fusion head enforces a structural reconstruction constraint that stabilizes cross-modal alignment and improves the interpretability of RGB–IR representations.
This head was designed with the following aspects: 1) The reconstruction constraint preserves texture, contrast, and edges in high-level fused features, preventing degeneration within GMIA and FSEC; 2) Sharing early-stage features with the segmentation head ensures that the network remains semantically discriminative while maintaining strong visual consistency.
Given the multi-scale features of RGB $\left\{ \mathbf{E}_1^{V}, \mathbf{E}_2^{V}, \mathbf{E}_3^{V}, \mathbf{E}_4^{V} \right\}$ and Infrared $\left\{ \mathbf{E}_1^{I}, \mathbf{E}_2^{I}, \mathbf{E}_3^{I}, \mathbf{E}_4^{I} \right\}$, these features are compressed via a Linear layer $\mathbf{F}_i^{m} = \phi_i(X_i^{m}),\quad m \in {{I},{V}}$. They are then upsampled $\tilde{\mathbf{F}}_i^{m} = \mathrm{Up}(\mathbf{F}_i^{m}, H, W).$
Finally, all multi-level features and the original inputs are concatenated and passed through shallow convolution to regress a fused image:
$\mathbf{Y}_{\text{fus}} = \sigma(
\mathrm{Conv}_{{fus}}([\tilde{\mathbf{F}}_i^{\text{I}},\tilde{\mathbf{F}}_i^{{I}}])
).$

\subsection{Loss Function}
To achieve the closed-loop optimization among enhancement, fusion, and segmentation, MRFS adopts a multi-task joint loss, consisting of a semantic segmentation loss and a fusion loss. The former ensures semantic accuracy and back-propagates semantic cues into early stages, while the latter enforces structural consistency across modalities.

\textit{Segmentation Loss:} Considering the class imbalance This paper adopt the Focal loss, which can be defined as follows:
\begin{align}
    \mathcal{L}_{\text{focal}}
= \alpha (1 - P(p))^{\gamma} \mathrm{CE}(p),
\end{align}
where $P(p)$ is predicted probability, $\alpha$ isbalancing factor, and $\gamma$ is focusing parameter.
To alleviate boundary blurriness induced by pixel-wise losses, IoU Loss is used:
\begin{align}
    \mathrm{IoU}_c
= \frac{\sum_p y_c(p)\hat{y}_c(p)}
{\sum_p y_c(p) + \sum_p \hat{y}_c(p) - \sum_p y_c(p)\hat{y}*c(p) + \epsilon },
\end{align}
The final semantic loss is:
\begin{align}
    \mathcal{L}_{\text{seg}}
= \mathcal{L}_{\text{focal}}+\mathcal{L}_{\text{IoU}}.
\end{align}

\textit{Fusion Loss:} 
The fusion loss enforces structural consistency of the fused image:
\begin{align}
    \mathcal{L}_{\text{fus}}
&= \beta_1 |I*{\text{fus}} - I_{\text{rgb}}|_1 + \beta_2 |I_{\text{fus}} - I_{\text{ir}}|_1 \\
&+ \beta_3 |\nabla I_{\text{fus}} - \nabla I_{\text{rgb/ir}}|_1,
\end{align}
where $I_{\text{fus}}$ denotes the reconstructed fused image produced by the fusion branch;
$I_{\text{rgb}}$ and $I_{\text{ir}}$ are the RGB and infrared inputs;
$\nabla(\cdot)$ is a gradient operator capturing edge and structural variations;
$\nabla I_{\text{rgb/ir}}$ represents the combined gradient constraints derived from both modalities;
$|\cdot|_1$ denotes the L1 norm;
and $\beta_1, \beta_2, \beta_3$ balance the contributions of intensity alignment, modality consistency, and structural preservation.
This loss provides an explicit reconstruction-driven regularization that enhances cross-modal consistency and supports the cooperative optimization of enhancement, fusion, and segmentation.

\textit{Overall Loss:} The final training objective is:
\begin{align}
    \mathcal{L}_{\text{total}} = \mathcal{L}_{\text{seg}} +  0.1*\mathcal{L}_{\text{fus}}.
\end{align}
This multi-task formulation empowers MRFS to achieve tightly coupled optimization across enhancement, fusion, and segmentation, leading to superior performance under diverse and challenging maritime conditions.





\section{EXPERIMENTS}
\subsection{Experimental Settings}
To the best of our knowledge, the constructed IVMSD dataset is the first infrared-visible ship multimodal dataset collected under real-world and complex maritime conditions.  To comprehensively validate the effectiveness of the proposed method, this paper have introduced additional infrared-visible datasets for validation, including the MFNet \cite{ha2017mfnet}, PST900 \cite{shivakumar2020pst900}, and FMB \cite{liu2023multi} datasets. Among these, MFNet primarily contains 1,569 infrared-visible image pairs from urban scenarios, making it suitable for autonomous driving research. PST900 focuses on underground environments, providing 1,038 calibrated infrared-visible image pairs dedicated to semantic segmentation of specific task-oriented targets. Meanwhile, FMB serves as a full-day multimodal benchmark comprising 1,500 infrared-visible image pairs, covering various adverse conditions in urban scenes.  
Although these datasets were not collected in maritime environments, the infrared-visible images from these diverse scenarios can sufficiently validate the cross-scene generalization capability and effectiveness of the proposed method.

\subsection{Implementation Details}
Considering the complex maritime conditions, to ensure information integrity and discriminability in both multimodal enhancement and downstream perception tasks, the feature encoder must possess strong representation capability. Thus, this paper adopts DINOv3 as the backbone for feature encoding. \textit{It is noted that the DINOv3's parameters are frozen during training}. The semantic segmentation and image fusion tasks are jointly trained for 200 epochs. The initial learning rate is set to ($1\times10^{-4}$) with a batch size of 4, optimized using the Adam optimizer. During training, the parameters of the feature encoder are frozen across all four datasets. Simultaneously, the image resolution is uniformly set to $480 \times 640$. During testing, the image resolutions for the IVMSD, MFNet, PST900, and FMB datasets are set to $720 \times 1280$, $480 \times 640$, $720 \times 1280$, and $600 \times 800$, respectively. All experiments are conducted on an NVIDIA GeForce RTX 4090 GPU with 24 GB of memory.


\subsection{Comparative Results and Analysis}
This paper first compares the proposed method with recent state-of-the-art approaches, including SeAFusion \cite{tang2022image}, SegForme \cite{xie2021segformer}, LASNet \cite{li2022rgb}, MDRNet+ \cite{zhao2023mitigating}, SGFNet \cite{wang2023sgfnet}, EAEFNet \cite{liang2023explicit}, MRFS \cite{zhang2024mrfs}, StitchFusion \cite{li2025stitchfusion}, and UniRGB-IR \cite{yuan2025unirgb}. It is worth noting that to comprehensively validate the effectiveness of the proposed method, experiments are conducted not only on the proposed IVMSD dataset but also on infrared-visible datasets from other scenarios (MFNet \cite{ha2017mfnet}, PST900 \cite{shivakumar2020pst900}, and FMB \cite{liu2023multi}) for cross-scenario verification.

On the IVMSD dataset, the proposed method achieves the best overall mIoU of 60.2\%, as reported in \autoref{IVMSD_RESULT_COMP}. It obtains strong IoU on large vessel categories, including cargo ships (59.3\%) and sand carriers (81.7\%), while remaining competitive on small targets such as speedboats (32.1\%) and fishing boats (28.9\%). In comparison, SeAFusion \cite{tang2022image}, LASNet \cite{li2022rgb}, and EAEFNet \cite{liang2023explicit} show noticeably lower performance on small-target categories, indicating that maritime degradations remain challenging for existing baselines.
To evaluate cross-scenario generalization, we further report results on MFNet, PST900, and FMB (\autoref{MFNet_result}, \autoref{PST900_result}, and \autoref{FMB_result}). The proposed method achieves the highest mIoU on all three datasets and outperforms the strongest competing method by 0.2, 2.6, and 4.8 percentage points on MFNet, PST900, and FMB, respectively. The larger margins on PST900 and FMB suggest that the framework is particularly effective under more challenging imaging conditions. We also observe that small-object categories remain comparatively difficult across methods, indicating that fine-grained boundary recovery and tiny-target representation are still open problems. Overall, these trends are consistent with the framework design, where FSEC improves degraded visual cues, SVCA enhances semantic consistency, and GMIA supports selective cross-modal interaction.

\begin{table}[t]
    \centering
    \caption{Quantitative experiments on the constructed IVMSD dataset}
    \resizebox{\linewidth}{!}{
        \begin{tabular}{c|cccc|c}
            \toprule
            \textbf{IVMSD}  & \begin{tabular}[c]{@{}c@{}}\textbf{Cargo} \\ \textbf{Ship}\end{tabular} & \textbf{\begin{tabular}[c]{@{}c@{}}Fishing\\ Boat\end{tabular}} & \textbf{\begin{tabular}[c]{@{}c@{}}Sand \\ Dredger\end{tabular}} & \textbf{Speedboat} & \textbf{mIoU}    \\ \midrule
            SeAFusion       & 21.1                & 0.0                   & 3.2                   & 0.0                & 24.6             \\
            SegFormer       & 47.0                & 24.5                  & 82.2                  & 22.3               & 55.0             \\
            LASNet          & 40.0                & 10.2                  & 49.2                  & 0.0                & 37.9             \\
            MDRNet+         & 23.2                & 0.01                  & 67.0                  & 10.1               & 40.4             \\
            SGFNet          & 28.5                & 0.1                   & 71.6                  & 12.7               & 43.6             \\
            EAEFNet         & 29.3                & 0.02                  & 60.1                  & 0.0                & 38.1             \\
            SegMiF          & 10.6                & 0.0                   & 5.1                   & 0.0                & 12.7             \\
            MRFS            & 41.0                & \underline{29.3}      & \textbf{83.8}         & \underline{28.5}   & 56.7             \\
            StitchFusion    & 41.7                & 13.1                  & 81.5                  & 15.7               & 50.2             \\
            UniRGB-IR       & 28.5                & 0.7                   & 64.8                  & 19.8               & 42.4             \\
            \midrule
            Ours (ViT-B)     & \underline{55.8}    & \textbf{30.4}         & 81.4                  & 25.7               & \underline{58.4} \\
            Ours (ViT-H)     & \textbf{59.3}       & 28.9                  & \underline{81.7}      & \textbf{32.1}      & \textbf{60.2}    \\ \bottomrule
        \end{tabular}
    }
    \label{IVMSD_RESULT_COMP}
\end{table}

\begin{table}[t!]
\centering
\caption{Quantitative experiments on the MFNet dataset.}
\resizebox{\linewidth}{!}{
\begin{tabular}{c|ccccc|c}
\toprule
\textbf{MFNet} & \textbf{Car} & \textbf{Person} & \textbf{Bike} & \textbf{Curve} & \textbf{Car Stop} & \textbf{mIoU} \\ \midrule
SeAFusion      & 84.2         & 71.1            & 58.7          & 33.1           & 20.1              & 48.8          \\
SegFormer      & 89.5         & 73.2            & 63.8          & 45.9           & 20.8              & 54.7          \\
EGFNet         & 87.6         & 69.8            & 58.8          & 42.8           & 33.8              & 54.8          \\
LASNet         & 84.2         & 67.1            & 56.9          & 41.1           & 39.6              & 54.9          \\
SegMiF         & 87.8         & 71.4            & 63.2          & 47.5           & 31.1              & 56.1          \\
MDRNet+        & 87.1         & 69.8            & 60.9          & 47.8           & 34.2              & 56.8          \\
SGFNet         & 88.4         & \textbf{77.6}  & 64.3          & 45.8           & 31.0              & 57.6          \\
MMSMCNet       & 89.2         & 69.1            & 63.5          & 46.4           & 41.9              & 58.1          \\
EAEFNet        & 87.6         & 72.6            & 63.8          & 48.6           & 35.0              & 58.9          \\
MRFS           & 89.4         & 75.4            & 65.0          & \textbf{49.0}  & 37.2              & 59.1          \\ \midrule
Ours           & \textbf{90.3} & 70.1           & \textbf{66.1} & 48.7          & \textbf{43.4}     & \textbf{59.3} \\ \bottomrule
\end{tabular}
}
\label{MFNet_result}
\end{table}

\begin{table}[t!]
\centering
\caption{Quantitative experiments on the PST900 dataset.}
\resizebox{\linewidth}{!}{
\begin{tabular}{c|cccc|c}
\toprule
\textbf{PST900} & \textbf{\begin{tabular}[c]{@{}c@{}}Hand-\\Drill\end{tabular}} & \textbf{BackPack} & \textbf{\begin{tabular}[c]{@{}c@{}}Frie-\\Extinguisher\end{tabular}} & \textbf{Survivor} & \textbf{mIoU} \\ \midrule
SeAFusion       & 65.6                & 59.6              & 41.1                       & 29.5              & 58.9          \\
SegFormer       & 74.3                & 86.4              & 61.1                       & 69.3              & 78.1          \\
EGFNet          & 64.7                & 83.1              & 71.3                       & 74.3              & 78.5          \\
LASNet          & 77.8                & 86.5              & 82.8                       & 75.5              & 84.4          \\
MDRNet+         & 63.0                & 76.3              & 63.5                       & 71.3              & 74.6          \\
SegMiF          & 66.0                & 81.4              & 76.3                       & 75.5              & 79.7          \\
MMSMCNet        & 62.4                & 89.2              & 73.3                       & 74.7              & 79.8          \\
SGFNet          & 82.8                & 75.8              & 79.9                       & 72.7              & 82.1          \\
EAEFNet         & 80.4                & 87.7              & 84.0                       & 76.2              & 85.6          \\
MRFS            & 79.7                & 87.4              & \textbf{88.0}                      & 79.6              & 86.9          \\ \midrule
Ours   & \textbf{83.3}                & \textbf{91.7}              & 87.3                       & \textbf{85.2}              & \textbf{89.5} \\ \bottomrule
\end{tabular}
}
\label{PST900_result}
\end{table}

\begin{table}[t!]
\centering
\caption{Quantitative experiments on the FMB dataset.}
\begin{tabular}{c|cccc|c}
\toprule
\textbf{FMB}   & \textbf{Car} & \textbf{Truck} & \textbf{T-Lamp} & \textbf{Buil.} & \textbf{mIoU} \\ \midrule
SeAFusion      & 76.2         & 15.1           & 34.4            & 80.1           & 51.9          \\
SegFormer      & 76.5         & 38.7           & 20.9            & 81.4           & 56.3          \\
LASNet         & 73.2         & 33.1           & 32.6            & 80.8           & 55.7          \\
SegMiF         & 78.7         & 42.4           & 35.6            & 80.1           & 58.5          \\
MDRNet+        & 75.4         & 27.0           & 41.4            & 79.8           & 55.5          \\
SGFNet         & 75.0         & 34.6           & 45.8            & 78.2           & 56.0          \\
EAEFNet        & 79.7         & 22.5           & 34.3            & 82.3           & 58.0          \\
MRFS           & 76.2         & 34.4           & 50.1            & \textbf{85.4 }          & 61.2          \\ \midrule
Ours  & \textbf{80.6}         & \textbf{72.1}           & \textbf{50.8}            & 84.7           & \textbf{66.0} \\ \bottomrule
\end{tabular}
\label{FMB_result}
\end{table}

\begin{table}[]
\centering
\caption{Ablation studies of the proposed modules.}
\resizebox{\linewidth}{!}{
    \begin{tabular}{cccc|cc}
    \toprule
    \multirow{2}{*}{Backbone}       & \multirow{2}{*}{FSEC}            & \multirow{2}{*}{GMIA}            & \multirow{2}{*}{SVCA}             & \multicolumn{2}{c}{mIoU} \\ \cmidrule(lr){5-5} \cmidrule(lr){6-6}
                                    &                                  &                                  &                                   & IVMSD       & MFNet      \\ \midrule
    \ding{51}                       & \textcolor{gray!70}{\ding{53}}   & \textcolor{gray!70}{\ding{53}}   & \textcolor{gray!70}{\ding{53}}    & 55.8        & 57.8       \\
    \ding{51}                       & \ding{51}                        & \textcolor{gray!70}{\ding{53}}   & \textcolor{gray!70}{\ding{53}}    & 57.4        & 58.5       \\
    \ding{51}                       & \ding{51}                        & \ding{51}                        & \textcolor{gray!70}{\ding{53}}    & 58.3        & 58.9       \\
    \ding{51}                       & \ding{51}                        & \ding{51}                        & \ding{51}                         & \textbf{60.2}        & \textbf{59.3}       \\ \bottomrule
    \end{tabular}
    }
    \label{ablation_study}
\end{table}

\subsection{Ablation Study}
\textit{1) Effectiveness of Each Proposed Module:}
The ablation results are summarized in \autoref{ablation_study}. Starting from the baseline backbone (55.8\% mIoU on IVMSD and 57.8\% on MFNet), progressively adding the proposed modules yields consistent improvements on both datasets. Adding FSEC increases mIoU by 1.6 percentage points on IVMSD and 0.7 percentage points on MFNet, indicating that frequency-spatial enhancement effectively mitigates degradations while preserving structural cues. Adding GMIA provides further gains of 0.9 and 0.4 percentage points, respectively, suggesting that gated cross-modal interaction improves feature complementarity and reduces redundant responses. The final addition of SVCA yields the largest incremental gain on IVMSD, which indicates that semantic-guided enhancement is particularly beneficial in challenging maritime conditions.

Overall, these results support the complementary roles of the three modules. FSEC operates as the restoration-oriented component and mainly addresses degradation-induced feature corruption by enhancing structural cues in both spatial and frequency representations, which improves the reliability of downstream perception. GMIA then serves as the fusion control component, selectively filtering cross-modal information to retain complementary responses while suppressing redundant or noisy interactions, thereby stabilizing multimodal feature aggregation. SVCA further acts as the semantic alignment component by feeding task-level semantic priors back to enhancement, encouraging the network to emphasize regions that are most relevant to segmentation rather than visually salient but semantically irrelevant patterns. The three modules couple restoration, cross-modal fusion, and semantic learning in a unified process. It indicates that task-guided joint optimization is more effective than generic fusion alone.

\begin{table}[t!]
    \centering
    \caption{Ablation study of different backbone on the IVMSD.}
    \resizebox{\linewidth}{!}{
        \begin{tabular}{c|cccc|c}
            \toprule
            \textbf{IVMSD}  & \begin{tabular}[c]{@{}c@{}}\textbf{Cargo} \\ \textbf{Ship}\end{tabular} & \textbf{\begin{tabular}[c]{@{}c@{}}Fishing\\ Boat\end{tabular}} & \textbf{\begin{tabular}[c]{@{}c@{}}Sand \\ Dredger\end{tabular}} & \textbf{Speedboat} & \textbf{mIoU}    \\ \midrule

            MRFS \cite{zhang2024mrfs}           & 41.0                & \underline{29.3}      & \textbf{83.8}         & {28.5}   & 56.7             \\
            AMDANet \cite{zhong2025amdanet}        & 44.7                & 28.4                  & 81.2                  & \underline{29.6}               & 56.6             \\
            \midrule
            Ours (mit-b4)    & \underline{56.1}    & {28.8}         & 80.2                  & 26.5               & {57.9} \\
            Ours (ViT-B)     & {55.8}    & \textbf{30.4}         & 81.4                  & 25.7               & \underline{58.4} \\
            Ours (ViT-H)     & \textbf{59.3}       & 28.9                  & \underline{81.7}      & \textbf{32.1}      & \textbf{60.2}    \\ \bottomrule
        \end{tabular}
    }
    \label{different_backbon}
\end{table}

\textit{2) Ablation Study of Different Backbone:} To ensure fairness, the framework is compared using an identical mit-b4 backbone, as shown in \autoref{different_backbon}. The proposed method achieves 57.9\% mIoU, surpassing MRFS and AMDANet at 56.7\% and 56.6\%, respectively. The 15.1\% margin in the Cargo Ship category over MRFS validates the specialized restoration design over generic attention. These gains are backbone-agnostic because scaling alone cannot resolve maritime-specific degradations like sea-fog or thermal noise. Consistent performance boosts across various encoders confirm that FSEC and SVCA address fundamental physical challenges that backbone strength alone cannot overcome. Consequently, the framework acts as an essential architectural complement to any encoder in complex maritime environments.


\textit{3) Ablation Study of the Hybrid Attention in the FSEC Module.} 
The hybrid attention mechanism is designed to dynamically weight and coordinate the feature information from both the frequency domain and the spatial domian, ensuring the constructive complementarity of the features. By ablating this mechanism, its necessity in achieving robust feature fusion is evaluated. As shown in \autoref{FSEC_ablation}, the table details the performance comparison on the IVMSD dataset when the hybrid attention mechanism is introduced and removed from the FSEC module. The experimental results clearly indicate that the hybrid attention is critical to the effectiveness of the FSEC module. Upon removing the hybrid attention, the model's mIoU sharply drops from 58.4\% to 53.9\%, resulting in a performance loss of up to 4.5\%. It is argued that the hybrid attention is not merely a means of performance enhancement, but rather a key success factor for the frequency-spatial enhancement and complementary strategy.

\begin{table}[t]
    \centering
    \caption{Ablation study of the Gated mechanism in FSEC, GMIA and SVCA module on the IVMSD dataset. The baseline is the DINOv3-Base model.}
    \resizebox{\linewidth}{!}{
        \begin{tabular}{c|c|cccc|c}
            \toprule
            \textbf{Module}         & \textbf{Gated}                 & \begin{tabular}[c]{@{}c@{}}\textbf{Cargo} \\ \textbf{Ship}\end{tabular} & \textbf{\begin{tabular}[c]{@{}c@{}}Fishing\\ Boat\end{tabular}} & \textbf{\begin{tabular}[c]{@{}c@{}}Sand \\ Dredger\end{tabular}} & \textbf{Speedboat} & \textbf{mIoU} \\ \midrule
            \multirow{2}{*}{{FSEC}} & \textcolor{gray!70}{\ding{53}} & \textbf{57.6}       & 29.5                  & 78.4                  & 24.8               & 57.8          \\
                                    & \ding{51}                      & 55.8                & \textbf{30.4}         & \textbf{84.4}         & \textbf{25.7}      & \textbf{58.4} \\ \midrule
            \multirow{2}{*}{{GMIA}} & \textcolor{gray!70}{\ding{53}} & \textbf{57.8}       & 26.4                  & 80.9                  & 24.3               & 57.7          \\
                                    & \ding{51}                      & 55.8                & \textbf{30.4}         & \textbf{84.4}         & \textbf{25.7}      & \textbf{58.4} \\ \midrule
            \multirow{2}{*}{{SVCA}} & \textcolor{gray!70}{\ding{53}} & \textbf{56.9}       & 27.0                  & 79.3                  & 25.6               & 57.5          \\
                                    & \ding{51}                      & 55.8                & \textbf{30.4}         & \textbf{84.4}         & \textbf{25.7}      & \textbf{58.4} \\ \bottomrule
        \end{tabular}
    }
    \label{tab:ablation_gated}
\end{table}

\textit{4) Ablation Study of the Gated Mechanism in Each Module.}
Furthermore, the gated mechanism is introduced across the three modules, and its performance gain is also presented in \autoref{tab:ablation_gated}. The gated mechanism is designed to achieve a dynamic and selective integration of cross-modal information, and its core function is to suppress the propagation of cross-modal redundancy and noise. In multi-modal fusion, the gated mechanism suppresses interference from low-quality or irrelevant modalities by dynamically weighting the contribution of different modal features. This ensures that the feature fusion process is guided by semantic consistency, thereby significantly improving the comprehensive performance across multiple modules.

\begin{table}[t]
    \centering
    \caption{Ablation study of the Hybrid Attention in the FSEC module on the IVMSD dataset}
    \resizebox{\linewidth}{!}{
        \begin{tabular}{c|cccc|c}
            \toprule
            \textbf{FSEC}   & \begin{tabular}[c]{@{}c@{}}\textbf{Cargo} \\ \textbf{Ship}\end{tabular} & \textbf{\begin{tabular}[c]{@{}c@{}}Fishing\\ Boat\end{tabular}} & \textbf{\begin{tabular}[c]{@{}c@{}}Sand \\ Dredger\end{tabular}} & \textbf{Speedboat} & \textbf{mIoU} \\
            \midrule
            w/o Hybrid Attn. & 51.7                & 19.0                  & 79.2                  & 20.7               & 53.9          \\
            w Hybrid Attn.  & \textbf{55.8}       & \textbf{30.4}         & \textbf{84.4}         & \textbf{25.7}      & \textbf{58.4} \\
            \bottomrule
        \end{tabular}
    }
    \label{FSEC_ablation}
\end{table}

\begin{figure*}[t!]
    \centering
    \includegraphics[width=\linewidth]{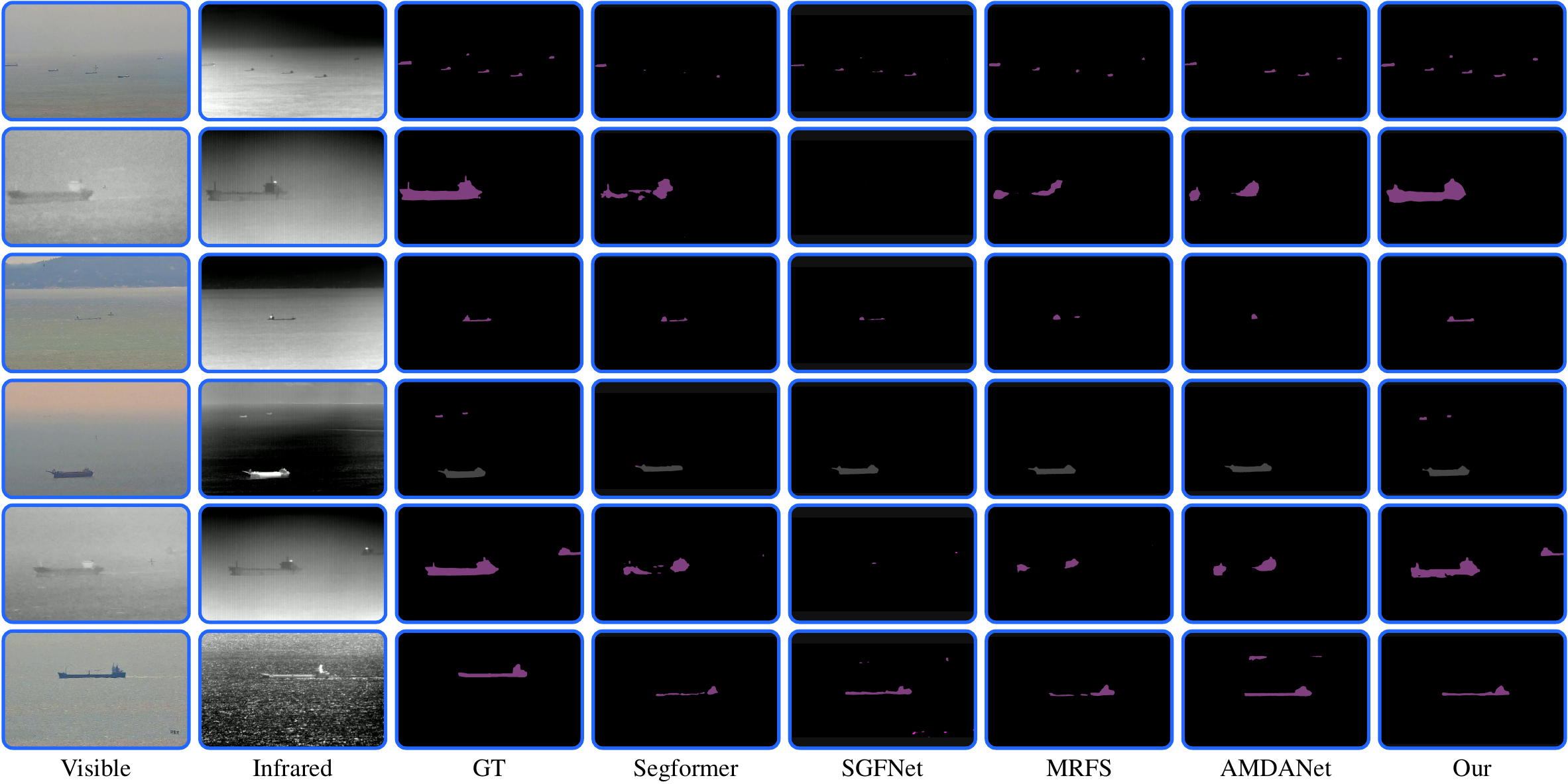}
    \caption{The visualization of the qualitative segmentation result.}
    \label{fig:visualization}
\end{figure*}


\textit{5) Ablation Study for the Proposed Module and Various Standard Techniques.}
\begin{table}[t!]
\centering
\caption{Ablation study for the proposed module and various standard techniques on the IVMSD.}
\begin{tabular}{lcc}
\toprule
\textbf{Model}                     & \textbf{Mechanism}      & \textbf{mIoU (\%)} \\ \midrule
\multirow{3}{*}{DINOV3 (ViT-Base)} & w Simple Fusion         & 53.9               \\
                                   & w Cross-Attn            & 54.6               \\
                                   & w GMIA                  & \textbf{55.1}               \\ \cmidrule{1-3} 
\multirow{2}{*}{DINOV3 (ViT-Base)} & w Frequency Process     & 54.6               \\
                                   & Ours                    & \textbf{56.0}               \\ \cmidrule{1-3} 
\multirow{2}{*}{DINOV3 (ViT-Base)} & w Self-Attn             & 54.2               \\
                                   & SVCA                    & \textbf{55.8}               \\ \bottomrule
\end{tabular}
\label{tab}
\end{table}
To demonstrate that the improvements come from the proposed components rather than backbone strength, component impacts are isolated against standard alternatives. The backbone is the frozen DINOv3 encoder. The results are reported in \autoref{tab}.
Simple fusion yields 53.9\% mIoU, replacing cross-attention at 54.6\% with GMIA at 55.1\% validates the proposed gated filtering. FSEC at 56.0\% significantly outperforms standard frequency processing at 54.6\%, proving specialized restoration exceeds generic filtering for maritime degradation. SVCA at 55.8\% substantially beats standard self-attention at 54.2\%. These fixed-encoder comparisons prove backbone strength alone is insufficient for maritime perception. Gains stem from architectural innovations providing the structural and semantic modeling required for complex sea conditions.

\begin{figure}[t]
    \centering
    \includegraphics[width=\linewidth]{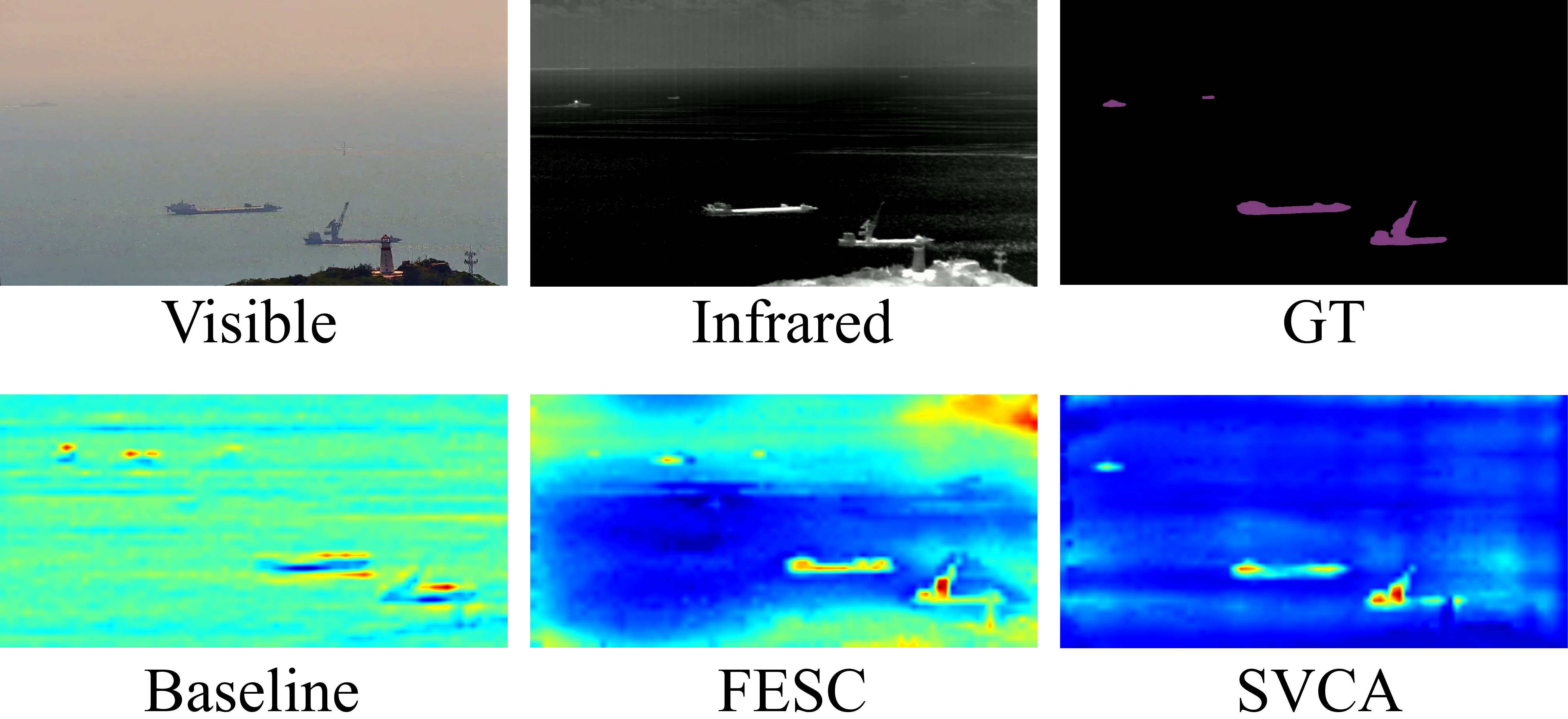}
    \caption{Feature Visualizations.}
    \label{fig:vis}
\end{figure}
\textit{6) Feature Visualization.}  
To demonstrate that the improved restoration yields better semantic understanding, feature visualizations are presented, as shown in \autoref{fig:vis}. It shows that the proposed modules drive semantic understanding. Baseline features are degraded by fog and reflections, exhibiting weak activations and surface noise. FSEC recovers ship structures and fine details, while SVCA suppresses background clutter and false positives via semantic priors. The shift from noisy baselines to sharp, localized activations proves that the proposed restoration mechanism reconstructs features for discriminability rather than mere aesthetics. This refinement correlates with mIoU gains in challenging scenes where standard encoders fail to isolate targets from environmental interference.

\subsection{Qualitative Analysis}
The qualitative visualization result is shown in \autoref{fig:visualization}. It clearly shows that the proposed method achieves the optimal segmentation performance. It can be attributes that the proposed FSEC mitigate degradation and recovers structural details and the SVCA provide semantic-consistent guidance for robust feature learning. In addition, GMIA enables selective and complementary fusion between infrared and visible modalities. These modules work together, enabling the proposed method to achieve the optimal performance. 

\section{CONCLUSION}
This paper addresses the challenges of semantic degradation and modality inconsistency in complex maritime environments. To support research on multimodal maritime perception, an Infrared–Visible Maritime Ship Dataset (IVMSD) is first constructed, which captures diverse illumination conditions, weather variations, and coupled degradations unique to marine scenes. Building on this dataset, the Multi-task Complementary Learning Framework (MCLF) is proposed to jointly integrate image restoration, multimodal fusion, and semantic segmentation within a unified architecture.
Extensive experiments demonstrate that the proposed method substantially improves segmentation robustness and perceptual quality under challenging maritime conditions, confirming the effectiveness of complementary learning across tasks and modalities. The constructed dataset and framework provide a strong foundation for future research on multimodal maritime scene and robust perception in degraded environments.

\bibliographystyle{IEEEtran}
\bibliography{IEEEabrv,reference}

\end{document}